\documentclass[11pt,a4paper]{article}
\usepackage[hyperref]{acl2021}
\usepackage{times}
\usepackage{latexsym}

\usepackage{graphicx}
\graphicspath{{figures/}}

\usepackage[inline,shortlabels]{enumitem}
\usepackage{amsmath}  %% <- after lineno
\usepackage{amsfonts}
\usepackage{microtype}

\aclfinalcopy % Uncomment this line for the final submission
\def\aclpaperid{1506} %  Enter the acl Paper ID here

\usepackage{flushend}

\usepackage{xcolor}
\usepackage[textsize=small]{todonotes}
\usepackage{ctable} %for the table (thicklines)
\usepackage{multirow}
\usepackage{fixltx2e}

\newcommand{\secref}[1]{\S\ref{#1}} % you can easily change to \S~\ref{#1} if you need to save some space ;-)

\newcommand{\figref}[1]{Fig.~\ref{#1}}    % within sentence
\newcommand{\Figref}[1]{Figure~\ref{#1}}  % start of sentence
\newcommand{\Tabref}[1]{Table~\ref{#1}}  % start of sentence
\newcommand{\equref}[1]{eq.~(\ref{#1})}

\newcommand{\eg}{e.g., }

\newcommand{\ie}{i.e., }

\newcommand{\dwiedataset}{DWIE}%{DW-Articles}
\newcommand{\docreddataset}{DocRED}%{DW-Articles}

\newcommand{\CEAFe}{CEAF$_\textrm{e}$}

\newcommand{\oldignore}[1]{}

\definecolor{RoyalBlue}{cmyk}{1, 0.50, 0, 0}
\newcommand{\revklim}[1]{#1}
\title{
\revklim{Injecting Knowledge Base Information into End-to-End Joint\\
Entity and Relation Extraction and Coreference Resolution}
}

\author{Severine Verlinden\textsuperscript{$*$}, Klim Zaporojets\textsuperscript{$*$}, Johannes Deleu, \\ \textbf{Thomas Demeester}, \textbf{Chris Develder} \\ 
  Ghent University – imec, IDLab \\
  Ghent, Belgium \\
\texttt{\{first\_name.last\_name\}@ugent.be} \\}

\date{}

\begin{document}
\maketitle
\begingroup\renewcommand\thefootnote{$*$}
\footnotetext{Equal contribution}
\endgroup
\begin{abstract}
%Many information extraction (IE) tasks often are addressed individually: named entity recognition (NER), relation
% We solve these tasks jointly, and over the whole document at once, using an entity-centric approach.
%%% ALT FOR above 2 lines:
We consider a joint information extraction (IE) model, solving named entity recognition, coreference resolution and relation extraction %(coref)
jointly over the whole document. %, using an entity-centric approach.
In particular, we study how to inject information from a knowledge base (KB) in such IE model, based on unsupervised entity linking. %, and demonstrate the performance increase realized through the KB addition.
The used KB entity representations are learned from either
(i)~hyperlinked text documents (Wikipedia), or
(ii)~a knowledge graph (Wikidata), and
appear complementary in raising IE performance.
%improve performance for the various tasks with an end-to-end system, solving all of NER, RE and coref in a single joint model. 
%%% This joint end-to-end IE setting goes beyond leveraging KBs for higher level language understanding
%%% %tasks (e.g., language understanding) %,  machine reading),  
%%% or specific individual IE tasks (e.g., entity typing). %, relation classification),
Representations of corresponding entity linking (EL) candidates are added %(derived from a heuristic string matching procedure)
to text span representations of the input document, and we experiment with  
% (i)~limiting this to a single candidate with highest prior (in Wikipedia), and 
% (i)~candidate prior (in Wikipedia) weights to combine the information in candidates, and
% (i)~weighting each of the EL candidates based on its prior (in Wikipedia), and
(i)~taking a weighted average of the EL candidate representations based on their prior (in Wikipedia), and
(ii)~using an attention scheme over the EL candidate list.
Results demonstrate an increase of up to 5\% F1-score for the evaluated IE tasks on two datasets.
Despite a strong performance of the prior-based model, our quantitative and qualitative analysis reveals the advantage of using the attention-based approach.

\end{abstract}

%------------------------------------------------------------------------------------------
\section{Introduction}
\label{sec:introduction}
%------------------------------------------------------------------------------------------
Information extraction (IE) comprises several subtasks, \eg named entity recognition (NER), coreference resolution (coref), relation extraction (RE). State-of-the-art results mainly report performance on single tasks, usually solving them on a sentence level (especially NER, RE).
However, in practice, IE system decisions should be consistent on the document level, \eg when processing news articles to automatically link entities (aside from potentially learning, \eg new relations).
Yet, the challenge of solving the tasks jointly on a document level has not received as much attention and remains hard \citep{durrett2014joint, yao2019docred, zaporojets2021dwie}.%\footnote{To evaluate such joint models, mention-level metrics based on whether individual mentions of entities are correctly extracted/typed/related to others, do not suffice to assess consistency across decisions aggregated over over the whole document (\cf entity-centric metrics proposed by \citet{zaporojets2020dwie}).}

On the other hand, it is well established that IE models benefit from incorporating background information of knowledge bases (KBs).
Still, so far this has been shown from the perspective of solving individual tasks such as relation classification or entity typing (\eg \citet{peters2019knowledge, liu2020k}).
Integrating KBs in joint models, realizing and analyzing the more complex end-to-end setting, has been left unexplored.

In terms of the nature of KBs adopted in IE, current approaches use either
\begin{enumerate*}[(i)]
\item structured knowledge \emph{graphs} comprising \texttt{(subj,rel,obj)} triples, \eg Wikidata \citep{yang2017leveraging,han2018neural,zhang2019long}, or
\item \emph{textual} descriptions, usually in hyperlinked  documents, \eg Wikipedia \citep{
% zhang2019ernie,
% ganea2017deep,
martins2019joint,
yamada2020luke}.
\end{enumerate*}
It has not been established to what extent KB-text and KB-graph 
% based 
entity
representations complement each other in boosting IE performance.

We address both research gaps of 
\begin{enumerate*}[(a)]
\item integrating KB information into a joint end-to-end IE model for solving named entity recognition, coreference resolution and relation extraction, and
\item analyzing what KB representation is more beneficial for IE, either 
  \emph{KB-graph} trained on Wikidata, or %\citep{joulin2017fast}, or % comprising \texttt{(subj,rel,obj)} triples, or
  \emph{KB-text} trained directly on Wikipedia.% \citep{yamada2016joint}. %on the Wikipedia hyperlinked text pages .
\end{enumerate*}
We particularly contribute:
\begin{enumerate*}[(i)]
\item a first span-based end-to-end architecture incorporating KB knowledge in a joint entity-centric setting, exploiting unsupervised entity linking (EL) to select KB entity candidates,
\item exploration of prior- and attention-based mechanisms to combine the EL candidate representations into the model,
\item assessment of the complementarity of KB-graph and KB-text representations, and
\item consistent gains of up to 5\% F1-score when incorporating KB knowledge in 3 document-level IE tasks evaluated on 2 different datasets. 
\end{enumerate*}

\section{Model}
\label{sec:model}
%------------------------------------------------------------------------------------------
% \begin{figure}[!ht]
% \centering
% \includegraphics[width=1.0\columnwidth, trim={0cm 0cm 0 0},clip]{figures/Overview_paper.png}
% \captionsetup{singlelinecheck=off}
% \caption[test]{TODO}
% \label{fig:model}
% \end{figure}
\begin{figure*}[ht]
\centering
\includegraphics[width=0.9\textwidth]{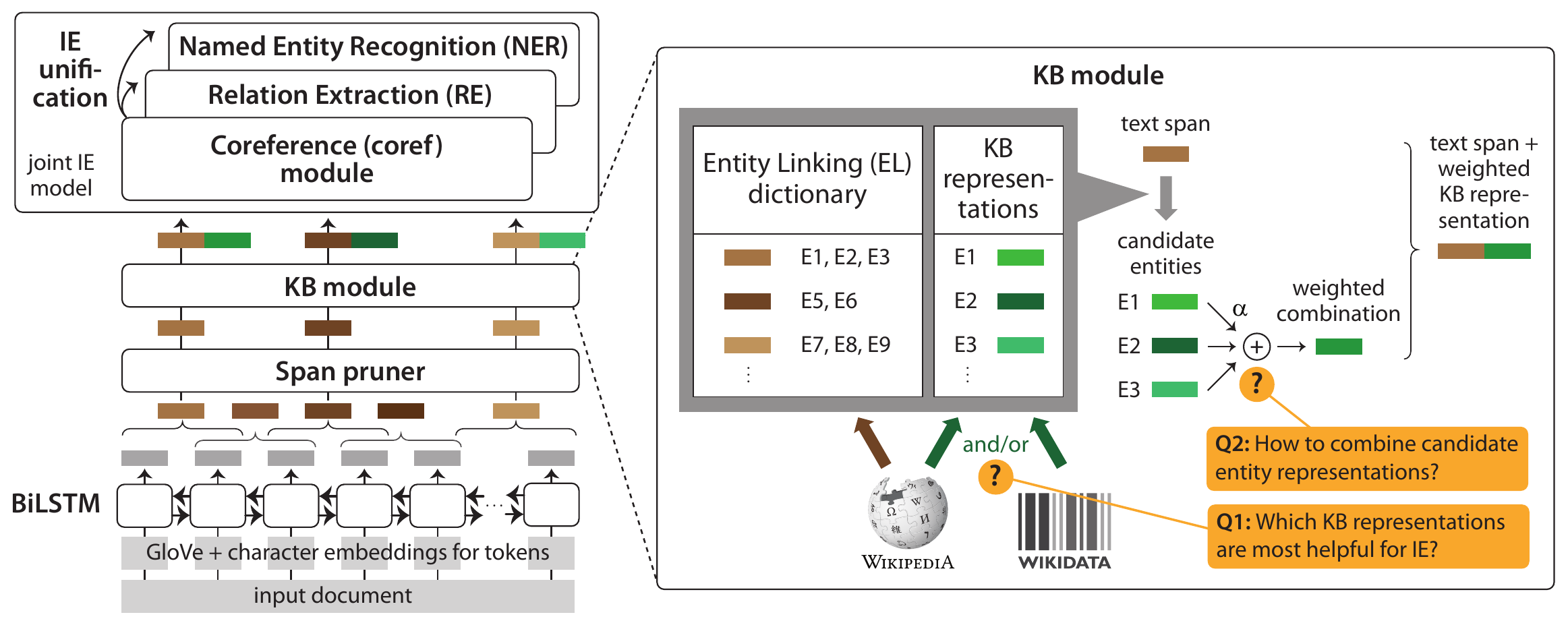}
\caption{Joint information extraction (IE) model with addition of a knowledge base (KB) module.}
\label{fig:model-overview}
\end{figure*}

\Figref{fig:model-overview} illustrates our model architecture.
Input document tokens are represented using concatenated GloVe \citep{pennington2014} and character embeddings \citep{ma2016} and pushed through a BiLSTM to obtain contextualized token representations, which are combined into spans.
Similar to \citet{luan2019general,zaporojets2021dwie}, a span pruner limits the number of spans for downstream modules.
The \emph{KB module} (\secref{sec:kb-module}) combines span representations with KB entity representations (\secref{sec:model-embeddings}), trained either on Wikidata (\emph{KB-graph}) or Wikipedia (\emph{KB-text}).
The KB-enriched span representations then serve as input for joint predictions on downstream IE tasks (\secref{sec:model-joint}). 

%------------------------------------------------------------------------------------------
% \subsection{Notations and Definitions}
% \label{sec:model-notations}
% to save space no "notations and definitions" for now unless we see it is necessary. 
%------------------------------------------------------------------------------------------

%------------------------------------------------------------------------------------------
\subsection{Entity Representations}
\label{sec:model-embeddings}

We experiment with 3 possible entity representations: \emph{KB-text}, \emph{KB-graph}, and concatenating \emph{both}. % (\emph{All}). 

\noindent\textbf{KB-text}: We follow \citet{yamada2016joint} 
% to train entity representations by leveraging the hyperlink structure of the Wikipedia pages. Concretely, the model is 
% trained 
to obtain the entity representations
using a skip-gram architecture \citep{mikolov2013efficient,mikolov2013distributed}, % jointly with entities by also
training to jointly predict
\begin{enumerate*}[(i)]
    \item the linked entities (through Wikipedia hyperlinks) given the target entity, and 
    \item the neighboring words for a given entity hyperlink. %a particular entity hyperlink. 
\end{enumerate*}
% instead of predicting the words, it is trained on predicting the correct entities based on surrounding words. 
% -> based on the hyperlink structure of wikipedia 

\noindent\textbf{KB-graph}: We adopt \citet{joulin2017fast} to train the entity embeddings directly on Wikidata triples \texttt{(subj,rel,obj)} by optimizing a linear classifier to predict the \texttt{obj} entity from the \texttt{subj} entity and the relation type \texttt{rel}. 

\oldignore{Unlike the KB entity embeddings, the KG representations are trained directly on the triples of Wikidata knowledge graph. We use the approach proposed by \citet{joulin2017fast} based on a linear classifier to predict the predicate entity according to the subject entity and the relation type. }
% We use the approach proposed by \citet{joulin2017fast} to train the entity embeddings directly on Wikidata KG triples by optimizing a linear classifier to predict the predicate entity from the subject entity and the relation type. 
% mention that we use this because of its simplicity?  
% , Instead, we frame this problem in a multiclass multilabel classification problem and model only the co-occurences of entities and relations
% with a linear classifier based on a Bag-of-Words (BoW) representation and standard cost functions.
% -> trained on triples without the text 
% \subsection{Candidate integration mechanisms}

%------------------------------------------------------------------------------------------
\subsection{KB module}
\label{sec:kb-module}
For a span $s_i$ from token $l$ to $r$, we obtain the 
% initial 
representation $\textbf{g}_i$
as input to the KB module
by concatenating the respective hidden LSTM states $\textbf{h}_l$ and $\textbf{h}_r$, and an embedding $\boldsymbol{\psi}_{r-l}$ for the corresponding span width $r-l$:
\begin{equation}
% \textbf{g}_i = [\textbf{h}_{l}; \textbf{h}_{r}; \boldsymbol{\psi}_{r-l}; \textbf{e}_i^x ] \label{eq:spanrepr} 
\textbf{g}_i = [\textbf{h}_{l}; \textbf{h}_{r}; \boldsymbol{\psi}_{r-l}]. \label{eq:spanrepr} 
\end{equation}
% For a given span $s_i$, we look up its candidate entities\footnote{We limit this to the 16 most frequent ones.} 
% $C_i$ in a dictionary built from Wikipedia, which also delivers for each candidate $c_{ij} \in C_i$ its prior probability $p_{ij}$ as per \citet[\S3]{yamada2016joint}.
We look up a given span $s_i$ in a dictionary built from Wikipedia, to determine its candidate entities set\footnote{We limit this to the 16 most frequent ones.} $C_i$, as well as the prior probability $p_{ij}$ for each $c_{ij} \in C_i$, as per \citet[\S3]{yamada2016joint}.

To combine the KB candidates $c_{ij}$, we either use
\begin{enumerate*}[(i)]
\item a uniform average (\emph{Uniform}),
\item the prior weights $p_{ij}$ (\emph{Prior}), 
\item an attention scheme (\emph{Attention}), or
\item attention with prior information (\emph{AttPrior}).
\end{enumerate*}
%%% We further define two attention schemas to weight the candidate entity $c_{i,j}$ of a particular span $s_i$ depending on whether the prior probability $p_{i,j}$ of the candidate is considered (\equref{eq:attprior}) or not (\equref{eq:att}),
% the information from the entity embeddings 
% depending on whether the weight comes 
% from attention (${A}$) or attention combined with prior (${J}$) as follows, % The attention-based ($A, J$) are predicted for each of the candidate entity $j \in C_i$ of span $s_i$ as follows,
The unnormalized 
attention scores for \emph{Attention} and \emph{AttPrior} are:
\begin{align}
 \Phi_\textit{Attention}(s_i,c_{ij},\textsc{k}) = \mathcal{F}_\textit{A} \left( [\textbf{g}_i; \boldsymbol{\xi}_\textsc{k}(c_{ij}) ]\right) \label{eq:att} \\
   \Phi_\textit{AttPrior}(s_i,c_{ij},\textsc{k}) = \mathcal{F}_\textit{AP} \big( [\textbf{g}_i; \boldsymbol{\xi}_\textsc{k}(c_{ij}); p_{ij} ]\big) \label{eq:attprior}   
\end{align}
where $\textsc{k} \in \{\emph{KB-text}, \emph{KB-graph}, \emph{both}\}$ refers to the entity representations from \secref{sec:model-embeddings}, $\boldsymbol{\xi}_\textsc{k}$ returns such representation for $c_{ij}$, and $\mathcal{F}_{*}$ is a feed-forward neural network (FFNN).
%, 
%\red{acting on the concatenation of span and entity representation, and for \textit{AttPrior} entity prior}. 
%
%\red{
%The entry representations for $c_{ij}$ are written $\boldsymbol{\xi}_\textsc{k}(c_{ij})$, where $\textsc{k} \in \{\emph{KB-text}, \emph{KB-graph}, \emph{both}\}$ refers to the entity representations from \secref{sec:model-embeddings}. 
%}
% and $p_{i,j}$ is the prior weight of the candidate $c_{i,j} \in C_i$ for a particular span $s_i$. These scores are used to calculate the normalized attention weights 
% using the softmax function. 
% Additionally, we experiment with prior weights that we refer as $\alpha^{\mathrm{prior}}_{i,j}$. 
% indicates the weighting mechanism defined in equations x,y and z respectively. 
The KB representation for span $s_i$ is a weighted average of its candidates $C_i$:
\begin{equation}
    % \textbf{e}_i^{x,y} = \sum_{j = 1}^{|C_i|} \alpha_{i,j}^{x,y} \cdot \boldsymbol{\xi}^{y}(c_{i,j}) \label{eq:end_emb}
    \textbf{e}_i^\textsc{k} = \sum_{c_{ij} \in C_i} %\sum_{j = 1}^{|C_i|}
    \mathcal{\alpha}_{ij} \cdot \boldsymbol{\xi}_\textsc{k}(c_{ij}) \label{eq:end_emb}     
\end{equation}
% Where $x \in \{A, J\}$. 
% The corresponding weighted average is also performed using the prior weights $p_{i,j}$.
where weights $\mathcal{\alpha}_{ij}$ either are uniform ($1/\left|C_i\right|$), the prior $p_{ij}$, or softmax-normalized attention scores (softmax over $\Phi$ from \equref{eq:att} or \equref{eq:attprior}).
The concatenation $[\textbf{g}_i; \textbf{e}_i^\textsc{k}]$ forms the KB-enriched representation for span $s_i$, as input for IE modules (\secref{sec:model-joint}).
%to use in downstream IE tasks discussed next (\secref{sec:model-joint}).  

%------------------------------------------------------------------------------------------
\subsection{Joint IE model}
\label{sec:model-joint}

% TODO - here briefly describe how relation extraction, coreference and NER are used. 
The joint IE model comprises 3 modules (\figref{fig:model-overview}) using the same KB-enriched representations $[\textbf{g}_i; \textbf{e}_i^\textsc{k}]$, and using a weighted combination of the 3 module losses to minimize during training. Note that NER and RE are framed as multi-label classification. % tasks. 

\noindent\textbf{NER module}: We use a FFNN on each span $s_i$ to produce scores $\boldsymbol{\Phi}_{\textsc{ner}} (s_i) \in \mathbb{R}^{|L_E|}$, with $L_E$ the set of possible entity types.
At inference, we accept type $l \in L_E$ for span $s_i$ if $\boldsymbol{\Phi}_{\textsc{ner}}(s_i)_l>0$. 

\noindent\textbf{Coref module}: We use the coreference scheme proposed by \citet{lee2017end}, using a FFNN to produce scores $\Phi_{\mathrm{coref}} (s_i, s_j)$: at inference time, the highest scoring antecedent of span $s_j$ is then chosen (potentially $s_j$ itself). %, in case of a singleton). %(including $s_j$ itself in case of singletons or non-entity spans).
% To account for singleton entity mentions, % in the %{\dwiedataset} and {\docreddataset} datasets,
Indeed, to allow for singletons we accept self-references $(s_j, s_j)$ if NER predicts the span $s_j$ to be an entity.

\noindent\textbf{RE module}: Similar to \citet{luan2019general,luan2018multi}, we use a FFNN to produce scores  $\boldsymbol{\Phi}_{\textsc{re}} (s_i, s_j) \in \mathbb{R}^{|L_R|}$ for each pair of spans $(s_i,s_j)$, with $L_R$ the set of relation types.
%We perform multi-label prediction and 
We accept relation $l \in L_R$ for pair $(s_i,s_j)$ if $\boldsymbol{\Phi}_{\textsc{re}} (s_i, s_j)_l > 0$. 

\noindent\textbf{IE unification}: Above modules make span level predictions. We obtain entity-centric predictions using the coref clusters, by assigning the union of predicted entity/relation types within a coref cluster to all its members, as do \citet{zaporojets2021dwie}. 

%Our joint loss is the weighted sum of the losses of each of the modules explained above. 

%------------------------------------------------------------------------------------------
\section{Experimental Setup}
\label{sec:experimental_setup}
%------------------------------------------------------------------------------------------

\begin{table}[t]
    \centering
    \resizebox{1.0\columnwidth}{!}
    %{\begin{tabular}{p{1cm} p{1.2cm} p{1.2cm} p{1.4cm} p{1.4cm}}
    {\begin{tabular}{lcccc}
        \toprule
         % Dataset & \begin{tabular}{@{}c@{}}\# Entity \\ clusters\end{tabular} & \begin{tabular}{@{}c@{}}\# Entity \\ types\end{tabular} & \# Relations & \begin{tabular}{@{}c@{}}\# Relation \\ types\end{tabular}\\
         \multirow{2}{*}{Dataset} & \# Entity & \# Entity & \multirow{2}{*}{\# Relations} & \# Relation \\
          & clusters & types &  & types \\         
         \midrule
         \dwiedataset & 23,130 & 311 & 21,749 & 65 \\
         \docreddataset & 98,610 & 6 & 50,503 & 96 \\
    \bottomrule
    \end{tabular}}
    \caption{Dataset statistics.} % for \dwiedataset~and \docreddataset.}
    \label{tab:dataset}
\end{table}

We evaluate our proposed models\footnote{\revklim{Code and models available at \url{https://github.com/klimzaporojets/e2e-kb-ie}.}} on  entity-centric multi-task datasets, summarized in \Tabref{tab:dataset}: {\dwiedataset} \citep{zaporojets2021dwie} and {\docreddataset} \citep{yao2019docred}. 
% \noindent\textbf{DWIE} Introduced by \citet{zaporojets2020dwie} this manually annotated entity-centric dataset contains . Furthermore, it contains manually annotated links, something that allows to 
% \noindent\textbf{DocRED} ... 
We report on coreference resolution (coref), NER and relation extraction (RE).
For coref, we report the average of 3 common F1 scores, as implemented by \citet{pradhan2014scoring}:
MUC~\citep{vilain1995model}, B$^3$~\citep{bagga1998algorithms} and {\CEAFe} \citep{luo2005coreference}.
Since we focus on entity-centric, document-level IE, for NER and RE we use \emph{hard} metrics \cite{zaporojets2021dwie} on the level of entity clusters (\ie aforementioned coref clusters): predictions are counted as correct only if
\begin{enumerate*}[(i)]
    \item all mentions (with exact boundary match) are present in the entity cluster, and 
    \item the predicted entity type (for NER) or relation type between two clusters (for RE) is correct.
\end{enumerate*}

Our experiments address 2 main questions (see \figref{fig:model-overview}):
\begin{enumerate*}[label=\textbf{(Q\arabic*)}]
   \item \label{it:q-kb-type} Which type of KB representation is most helpful for IE (\emph{KB-text}, \emph{KB-graph}, or \emph{both}; see \secref{sec:model-embeddings})?
    \item \label{it:q-attention} Which weighting scheme to use for $\alpha$ (\emph{Uniform}, \emph{Prior}, \emph{Attention}, \emph{AttPrior}; see \secref{sec:kb-module})?
\end{enumerate*}

% \todo[inline]{KZ: nnet details such as learning rate, numbers of layers in FFNN networks, iterations, dropout, entity embedding dimensionality, etc. TD: not possible in short paper format; are most of these kept from DWIE? If so, maybe refer to that, and mention the KB representation sizes? Also, here would be the place to refer to the code online (still anonymous).}
% requires

%------------------------------------------------------------------------------------------
\section{Results}
\label{sec:results}
%------------------------------------------------------------------------------------------

\begin{table*}
\centering
\resizebox{0.8\textwidth}{!}{\begin{tabular}{ccclllllll}
\toprule
 &  & \multicolumn{3}{c}{\dwiedataset} && \multicolumn{3}{c}{\docreddataset}\\
\cmidrule(lr){3-5}\cmidrule(lr){7-9}
 KB Source & Setup & \multicolumn{1}{c}{Coref } &    \multicolumn{1}{c}{NER} & \multicolumn{1}{c}{RE} && \multicolumn{1}{c}{Coref } &    \multicolumn{1}{c}{NER} & \multicolumn{1}{c}{RE}\\
%  Setup & Emb. &  Avg.&  & $F_{1,e}$  &  & $F_{1,e}$ &&& Avg.&  & $F_{1,e}$  &  & $F_{1,e}$\\
\toprule
-- & Baseline & 90.0${\scriptstyle \pm\text{0.2}}$
% {\scriptstyle \pm0.2} 
& 71.7${\scriptstyle \pm\text{0.5}}$ & 47.0${\scriptstyle \pm \text{1.4}}$ && 81.9${\scriptstyle \pm \text{0.3}}$ & 68.5${\scriptstyle \pm \text{0.3}}$ & 23.5${\scriptstyle \pm \text{0.6}}$ \\
\midrule
 & Uniform & 90.7${\scriptstyle \pm\text{0.2}}$ & 73.5${\scriptstyle \pm \text{0.5}}$ & 48.5${\scriptstyle \pm \text{1.1}}$ && 82.9${\scriptstyle \pm \text{0.1}}$ & 70.7${\scriptstyle \pm \text{0.2}}$ & 24.5${\scriptstyle \pm \text{0.3}}$\\
KB-text  & Attention&  \textbf{90.7${\scriptstyle \pm \text{0.3}}$} & 73.4${\scriptstyle \pm \text{0.8}} $ & 49.0${\scriptstyle \pm \text{0.4}}$ && \textbf{83.4${\scriptstyle \pm \text{0.1}}$} & 71.2${\scriptstyle \pm \text{0.1}}$ & 24.5${\scriptstyle \pm \text{0.3}}$\\
 & AttPrior& 90.7${\scriptstyle \pm\text{0.3}} $ & 73.7${\scriptstyle \pm\text{0.6}}$ & \textbf{49.6${\scriptstyle \pm\text{0.8}}$} && 83.2${\scriptstyle \pm\text{0.2}}$ & \textbf{71.3${\scriptstyle \pm\text{0.2}}$} & 24.8${\scriptstyle \pm\text{0.4}}$\\
 & Prior& 90.7${\scriptstyle \pm\text{0.2}}$ & \textbf{73.8${\scriptstyle \pm\text{0.5}}$} & 49.4${\scriptstyle \pm\text{0.4}}$ && 82.9${\scriptstyle \pm\text{0.2}}$ & 70.9${\scriptstyle \pm\text{0.3}} $ & \textbf{25.3${\scriptstyle \pm\text{0.4}}$}\\
%% & Oracle (links) & $ 93.2{\scriptstyle \pm0.2} $ & $79.5{\scriptstyle \pm0.4} $ & $57.0{\scriptstyle \pm0.8}$\\
\midrule
  & Uniform & 91.0${\scriptstyle \pm\text{0.3}} $ & 73.6${\scriptstyle \pm\text{0.4}} $ &  48.0${\scriptstyle \pm\text{1.2}}$ && 83.3${\scriptstyle \pm\text{0.2}} $ &  71.1${\scriptstyle \pm\text{0.2}} $ & 24.9${\scriptstyle \pm\text{0.2}}$\\
KB-graph & Attention & 91.2${\scriptstyle \pm\text{0.3}}$ & 73.9${\scriptstyle \pm\text{0.5}}$ & 50.1${\scriptstyle \pm\text{1.1}}$ && \underline{\textbf{83.7${\scriptstyle \pm\text{0.1}}$}} & \textbf{71.6${\scriptstyle \pm\text{0.1}}$} & 25.0${\scriptstyle \pm\text{0.4}}$\\
  & AttPrior & \textbf{91.3${\scriptstyle \pm\text{0.2}}$} & \textbf{74.6${\scriptstyle \pm\text{0.3}}$} & \textbf{50.5${\scriptstyle \pm\text{1.0}}$} && 83.5${\scriptstyle \pm\text{0.3}}$ & 71.5${\scriptstyle \pm\text{0.2}}$ & 25.1${\scriptstyle \pm\text{0.2}}$\\
  & Prior & 90.8${\scriptstyle \pm\text{0.3}}$ & 73.6${\scriptstyle \pm\text{0.6}}$ & 49.6${\scriptstyle \pm\text{1.1}}$ && 83.4${\scriptstyle \pm\text{0.1}}$ & 71.1${\scriptstyle \pm\text{0.1}}$ & \textbf{25.2${\scriptstyle \pm\text{0.2}}$}\\
%% & Oracle (links) & 93.2${\scriptstyle \pm \text{0.2}}$ & 79.8${\scriptstyle \pm \text{0.4}}$ &  55.5${\scriptstyle \pm \text{1.7}}$\\
\midrule
both & Uniform & 91.1${\scriptstyle \pm\text{0.1}}$  & 74.1${\scriptstyle \pm\text{0.5}}$ & 49.3${\scriptstyle \pm\text{0.5}}$ && 83.5${\scriptstyle \pm\text{0.1}}$ & 71.3${\scriptstyle \pm\text{0.2}}$  & 24.8${\scriptstyle \pm\text{0.1}}$\\
(KB-graph + & Attention & 91.2${\scriptstyle \pm\text{0.3}} $ & 74.3${\scriptstyle \pm\text{0.6}}$ & 51.3${\scriptstyle \pm\text{1.3}}$ && 83.5${\scriptstyle \pm\text{0.2}}$ & 71.5${\scriptstyle \pm\text{0.1}}$ & 24.8${\scriptstyle \pm\text{0.3}}$\\
KB-text) & AttPrior & \underline{\textbf{91.5${\scriptstyle \pm\text{0.2}}$}} & \underline{\textbf{75.0${\scriptstyle \pm\text{0.4}}$}} & \underline{\textbf{52.1${\scriptstyle \pm\text{1.2}}$}} && \textbf{83.6${\scriptstyle \pm\text{0.2}}$} & \underline{\textbf{71.8${\scriptstyle \pm\text{0.3}}$}} & \underline{\textbf{25.7${\scriptstyle \pm\text{0.7}}$}}\\
  & Prior & 90.8${\scriptstyle \pm\text{0.1}}$ & 73.8${\scriptstyle \pm\text{0.2}}$ & 49.8${\scriptstyle \pm\text{1.2}}$ && 83.2${\scriptstyle \pm\text{0.1}}$ & 71.2${\scriptstyle \pm\text{0.1}}$ & 25.1${\scriptstyle \pm\text{0.3}}$\\
%% & Oracle (links) &  93.6${\scriptstyle \pm \text{0.3}}$ & 80.5${\scriptstyle \pm \text{0.3}}$ & 56.9${\scriptstyle \pm \text{1.0}}$\\
\bottomrule
\end{tabular}}
\caption{Main results of the experiments in F1 scores grouped by the background KB source. We report Avg.\ F1 scores of MUC, B$^3$ and {\CEAFe} for Coref, and hard F1 metrics for NER and RE. \textbf{Bold} font indicates the best results for each of the different \emph{KB source} types. Additionally, the best overall results are \underline{underlined}.}
\label{tab:overview_results}
\end{table*}

We summarize the comparison of various model choices for both {\dwiedataset} and {\docreddataset} datasets in \Tabref{tab:overview_results}.
First, looking into \ref{it:q-kb-type}, we note that including background information from \emph{KB-graph} and \emph{KB-text} significantly boosts performance compared to the \emph{Baseline} without any KB. \revklim{Additionally, our model outperforms the results from \citet{zaporojets2021dwie} (not listed in the table) by about 2 percentage points F1, using the same input (GloVe) representations.}
Furthermore, 
% best results are achieved by combining \emph{both}, 
\revklim{we observe a general improvement in results when combining \emph{both} representations,}
suggesting that a (hyper)text corpus (Wikipedia) and a knowledge graph (Wikidata) embed complementary information for raising IE performance. 

Deeper analysis reveals that adding KB representations mainly benefits performance for ``rare'' entity types: \eg limiting the test set to entity types that occur $\leq$50 times in the training set for {\dwiedataset}, compared to \emph{Baseline}, F1 for NER goes up by $+$13.9 for \emph{KB-both} with \emph{AttPrior}, while the benefit gradually decreases for more frequently occurring entity types.
%(\eg limiting the test set to entities that occur $<$50 times in the full corpus, compared to \emph{Baseline}, F1 goes up by $+$17.8 for \emph{KB-both} with \emph{AttPrior}).
For RE, we note that overall we also see a clear 
% overall 
performance gain from adding KB information (\eg +5.1\% F1 for \emph{both} KB sources with \emph{AttProp} compared to \emph{Baseline} for {\dwiedataset}), yet the boost is not as clear for relations with fewer training instances.
(The latter makes sense, since we inject KB representations of entities rather than explicitly also for relations; we leave studying adding relation embedding information for future work.)
\begin{figure}[!t]
\centering
\includegraphics[width=0.85\columnwidth, trim={0.2cm 10.2cm 16.4cm 0.2cm},clip]{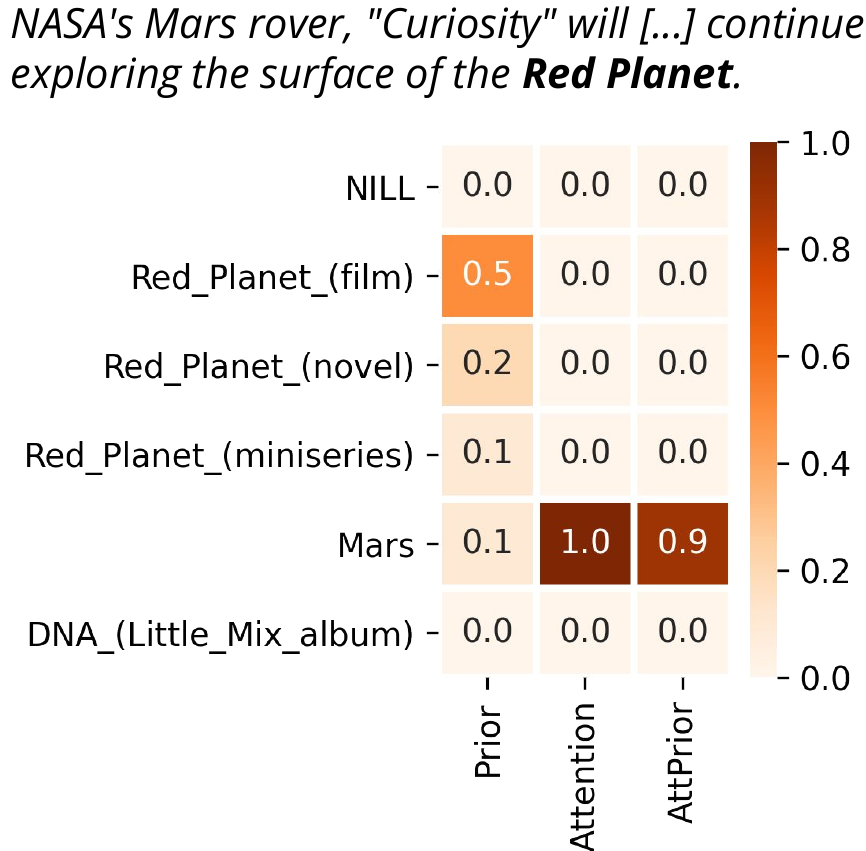}
\captionsetup{singlelinecheck=off}
\caption[test]{Illustration of EL candidate weighting: the $\alpha$ weights for top candidates for ``Red Planet'' from the example sentence at the top. 
Attention-based weighting (\emph{Attention}, \emph{AttPrior}) correctly identify the ``Mars'' entity, while the Wikipedia-based \emph{Prior} fails, as most of Wikipedia's ``Red Planet'' links refer to the film.}
%Example illustrating how attention-based \emph{Attention} and \emph{AttPrior} schemas correctly identify the entity \emph{Mars} from the candidate list as the most relevant. This is missed by the \emph{Prior} since most of the mentions of ``Red Planet'' co-occur with the \emph{Red\_Planet\_(film)} entity in Wikipedia corpus.}
\label{fig:concrete_example}
\end{figure}

Second, for \ref{it:q-attention}, we note that the \emph{AttPrior} scheme is the overall winner among the different EL candidate weigthing schemes. %(although in some KB setting and dataset combinations other non-uniform schemes , although for  of KB setting and the dataset \emph{Prior} and \eweighting of EL candidates ale
We observed that in terms of ranking EL candidates, \emph{Prior} performs quite well on {\dwiedataset} --- for 86.5\% of entity mentions it assigns the highest score to the correct EL candidate, while \emph{Attention} and \emph{AttPrior} achieve it for 46.2\%, resp.\ 77.2\% of the mentions --- which basically confirms that 
% the datasets 
{\dwiedataset}
has a similar entity distribution as Wikipedia.\footnote{{\dwiedataset} is a news article corpus.
% ; {\docreddataset} is actually a subset of Wikipedia.
}
Yet, it seems necessary to include alternative candidates, and the attention-based schemes thus can correct EL mistakes of \emph{Prior}, as illustrated in \figref{fig:concrete_example}.
This correction leads to a resulting boost for the IE tasks as reported in \Tabref{tab:overview_results}.
E.g., we found that for {\dwiedataset}, looking at clusters with entity mentions for which \emph{Prior} makes wrong EL predictions, the \emph{AttPrior} weighting scheme retrieves $+$3.7\% more of the gold standard annotated named entities (as opposed to just $+$0.6\% in the clusters with correct \emph{Prior} EL candidates).
Perfecting the EL prediction would potentially boost IE performance even more.

\section{Related Work}
\label{sec:related_work}
%------------------------------------------------------------------------------------------
As stated earlier, we studied how to integrate 
\begin{enumerate*}[(i)]
\item \label{it:kb-into-ie} knowledge base information into IE, and particularly
\item \label{it:e2e-ie} end-to-end IE combining multiple tasks (NER, relation extraction, coreference resolution), while 
\item \label{it:entity-centric-ie} taking an entity-centric perspective, \ie focus on making consistent decisions on the document level.
\end{enumerate*}
%We highlight relevant works for each of \ref{it:kb-into-ie}--\ref{it:entity-centric-ie} below.
%
For \ref{it:kb-into-ie}, integrating KB into IE has been applied for individual tasks: relation classification \citep{poerner2020bert, zhang2019long,yang2017leveraging}, entity typing \citep{peters2019knowledge} and NER \citep{yamada2020luke}.
For \ref{it:e2e-ie}, recently span-based architectures \citep{lee2017end,luan2019general,wadden2019entity,fei2020boundaries} have been proposed.
Our work unifies the KB integration concept into such span-based IE system, in particular an entity-centric one (as per \ref{it:entity-centric-ie}), building on \citet{jia2019document, zaporojets2021dwie}.
For the KB integration approach, we exploit entity representations trained on a hypertext corpus, as in \citep{yamada2016joint, % zhang2019ernie,
ganea2017deep, yamada2020luke} or learnt from a knowledge graph \citep{yang2017leveraging,han2018neural,zhang2019long}.
Our results show that both offer complementary value for IE. 
\revklim{Similarly to our work, \citet{yamada2019neural} also explore using an attention-weighted combination of entity representations, but they use it to build a full document representation (with mentions having the entities as candidates) for a text classification task. In contrast, our span-based attention model
% is applied on each of the entity mention candidates. This enables our model to 
is able to ``inject'' knowledge in each of the mentions separately, for more fine-grained downstream IE tasks that are mention-dependent, \eg coreference resolution, relation extraction and NER. }
% \citet{yamada2019neural}

%------------------------------------------------------------------------------------------
\section{Conclusion}
\label{sec:conclusion}
%------------------------------------------------------------------------------------------
We propose an end-to-end model for joint IE (NER + relation extraction + coreference resolution) incorporating entity representations from a background knowledge base (KB), using a span-based system.
We find that representations built from a knowledge graph and a hypertext corpus are complementary in boosting IE performance.
To combine candidate entity representations for text spans, we explore various weighting schemes: an attention-based combination is successful in combining prior frequency information from a hypertext corpus with contextual information to identify the relevant entity, and achieves highest IE performance. % relevant for a span.

\oldignore{
\textbf{-----------OLD TEXT-----------}\\
In this paper, we propose to use knowledge graph and hyperlinked corpus-based entity representations to inject background knowledge in a 
% span-based
joint 
% entity-centric
end-to-end IE framework. 
We find that both of the information sources are complementary and give superior results when used together. 
Additionally, we explore the prior and attention weighting schemes to combine the candidate entity representations for each of the spans. 
Our experiments show the advantage of using attention-based weights, specially in the corner cases where the entity mention is used to refer to less frequently co-occurring entities.
% In this paper, we explore prior and attention-based techniques to incorporate background knowledge using entity representations in a joint end-to-end entity-centric  IE setting. 
% Our experiments show a significant improvement on all of the evaluated tasks when ... This improvement ... 
}

\section*{Acknowledgments}
% ==============================================================================
\revklim{\noindent Part of the research leading to these results has received funding from
\begin{enumerate*}[(i)]
\item the European Union's Horizon 2020 research and innovation programme under grant agreement no.\ 761488 for the CPN project,\footnote{\url{https://www.projectcpn.eu/}} and
\item the Flemish Government under the programme ``Onderzoeksprogramma Artifici\"{e}le Intelligentie (AI) Vlaanderen''.
\end{enumerate*}
}
% \bibliography{anthology,acl2020}

\bibliography{bibliography}
\bibliographystyle{acl_natbib}

%% end oldignore
\end{document}